\documentclass[10pt,twocolumn,letterpaper]{article}
\newcommand{\eg}{\textit{e.g.}}

\newcommand{\ie}{\emph{i.e.}}

\newcommand{\tb}{\textbf}

\newcommand{\ti}{\textit}

\newcommand{\bs}{\boldsymbol}

\newcommand{\mb}{\mathbb}

\newcommand{\rr}{\mb{R}}

\usepackage{algorithm}
\usepackage{algorithmicx}

\usepackage{cvpr}              % To produce the CAMERA-READY version
\usepackage[accsupp]{axessibility}
\usepackage[T1]{fontenc}
\usepackage{graphicx}
\usepackage{multirow}
\usepackage{amsmath}
\usepackage{amssymb}
\usepackage{booktabs}

\usepackage[pagebackref,breaklinks,colorlinks]{hyperref}

\usepackage[capitalize]{cleveref}
\crefname{section}{Sec.}{Secs.}
\Crefname{section}{Section}{Sections}
\Crefname{table}{Table}{Tables}
\crefname{table}{Tab.}{Tabs.}

\def\cvprPaperID{11197} % *** Enter the CVPR Paper ID here
\def\confName{CVPR}
\def\confYear{2024}

\title{
\vspace{-2em}
Multimodal Sense-Informed Forecasting of 3D Human Motions}
\author{
  Zhenyu Lou\textsuperscript{\rm 1} \,\,\,\,\,\, Qiongjie Cui\textsuperscript{\rm 2}\thanks{Corresponding author}\\  
  Haofan Wang\textsuperscript{\rm 3} \,\,\,\,\,\, Xu Tang\textsuperscript{\rm 3} \,\,\,\,\,\, 
  Hong Zhou\textsuperscript{\rm 1} \\
  \textsuperscript{\rm 1}Zhejiang University, 
  \textsuperscript{\rm 2}Nanjing University of Science and Technology, 
  \textsuperscript{\rm 3}Xiaohongshu Inc
  \\
  {\tt\small
  11915044@zju.edu.cn\,\,\,\,  cuiqiongjie@126.com
  }
}

\begin{document}

\makeatletter
\let\@oldmaketitle\@maketitle% Store \@maketitle
\renewcommand{\@maketitle}{\@oldmaketitle% Update \@maketitle to insert...
	\captionsetup{type=figure}
    \centering
    \includegraphics[width=0.99\linewidth]{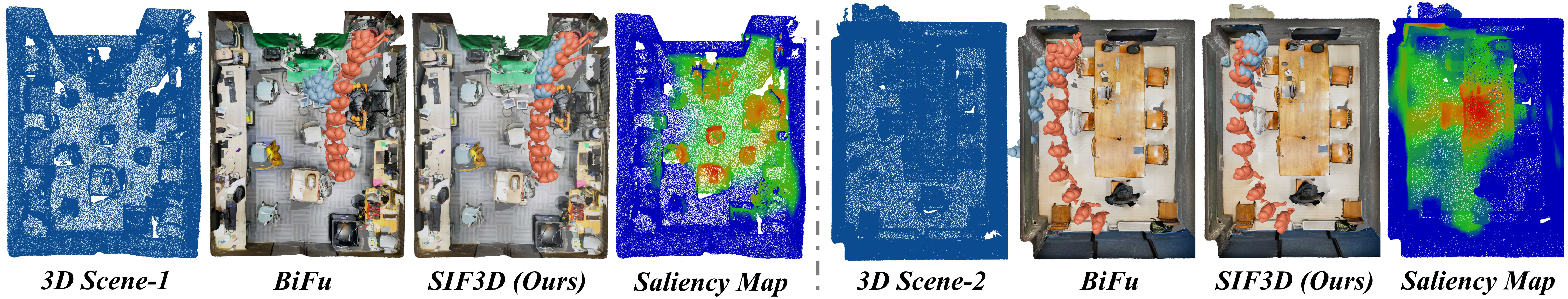}
	\captionof{figure}{\ti{\tb{The proposed SIF3D:} multimodal \tb{S}ense-\tb{I}nformed \tb{F}orecasting of \tb{3D} human motions.
        Our SIF3D takes the observed motion sequence, as well as the 3D scene point cloud as input modalities, and is able to identify salient points (redder) and underlying ones (bluer), to generate the accurate trajectory and high-fidelity future poses within given 3D scenarios.
        In contrast, the state-of-the-art baseline of BiFu \cite{zheng2022gimo} equally considers the global scene embedding, and thus cannot distinguish the saliency of the 3D scene, leading to the physically implausible motions, e.g., human mesh intersecting or distorting with the 3D environment, violating any physical constraints.
        }}
	\label{f1}
	\bigskip}
\makeatother

\maketitle

\begin{abstract}
    \vspace{-.5em}
    Predicting future human pose is a fundamental application for machine intelligence, which drives robots to plan their behavior and paths ahead of time to seamlessly accomplish human-robot collaboration in real-world 3D scenarios.
    Despite encouraging results, existing approaches rarely consider the effects of the external scene on the motion sequence, leading to pronounced artifacts and physical implausibilities in the predictions.
    To address this limitation, this work introduces a novel multi-modal sense-informed motion prediction approach, which conditions high-fidelity generation on two modal information: external 3D scene, and internal human gaze, and is able to recognize their salience for future human activity.
    Furthermore, the gaze information is regarded as the human intention, and combined with both motion and scene features, we construct a ternary intention-aware attention to supervise the generation to match where the human wants to reach.
    Meanwhile, we introduce semantic coherence-aware attention to explicitly distinguish the salient point clouds and the underlying ones, to ensure a reasonable interaction of the generated sequence with the 3D scene.
    On two real-world benchmarks, the proposed method achieves state-of-the-art performance both in 3D human pose and trajectory prediction.
    More detailed results are available on the page: 
    \url{https://sites.google.com/view/cvpr2024sif3d}.
\end{abstract}

\vspace{-1.0em}
\section{Introduction}
\vspace{-0.2em}
\label{sec:1}
Forecasting future human poses, from the observed ones, stands as a fundamental application in the domain of machine intelligence, autonomous vehicles, and human-robot collaboration
\cite{Gui2018AdversarialGH,mao2019learning,wang2021pvred,liu2021aggregated,dang2021msr,piergiovanni2020adversarial,zheng2022gimo,ma2022progressively,li2022skeleton,diller2022forecasting,xu2023auxiliary}.

Human motion is fundamentally intertwined with and constrained by its environment \cite{su2022crossmodal, liu2021multimodal, radwan2020multimodal},
a reality often overlooked in previous motion prediction methods \cite{mao2019learning, aksan2021spatio, ma2022progy, li2022skeleton}. 
This oversight could lead to noticeable anomalies and unrealistic outcomes in the predictions.
We notice this issue and aim to solve it.

In robotics, scene information is commonly represented as a point cloud \cite{chen2019deep, liu2019flownet3d, rusu2009close, yang2016automatic}, where a large number of points are constructed as a collection to describe a 3D environment.
Existing methods perform motion prediction/generation by encoding the entire scene information into a unified global embedding \cite{cao2020long, hassan2021stochastic, zheng2022gimo}.
Despite their good performance, we note that not all information within the point cloud is equally relevant to the motion prediction task; instead, only a small subset is salient.
Moreover, the global embedding fails to capture the intricate local details of the scene; therefore, the generation may deviate from the intended scene semantics.
In addition, the human gaze is a valuable manifestation, providing an intention that the human is likely to reach or a direction that the human is likely to move toward \cite{ghosh2023imos, trick2019multimodal, yu2015human}.
By jointly considering the 3D scene and human gaze, it is possible to capture the human's behavior of wanting to move toward a specific location, and thus generate more reasonable motion sequences.

Towards this end, we propose a novel multimodal sense-informed predictor for real-world 3D scenarios, named SIF3D.
It integrates past motion sequences with critical data from external 3D scenes and the observer's internal gaze to enhance 
scene-aware human activity representations \cite{corona2020context, cao2020long, hassan2021stochastic, zheng2022gimo}.
Following the recent progress, both motion encoder and scene encoder are introduced to extract the motion embedding and scene embedding \cite{qi2017pointnet++}, respectively.
To further refine point cloud data utilization and focus on key points affecting actions, 
we propose two novel scene-aware cross-modal attention mechanisms: 
ternary intention-aware attention (TIA) integrates human gaze, motion, and scene data through a comprehensive analysis to concentrate on globally salient points;
semantic coherence-aware attention (SCA) focuses on identifying the local salient scene points by analyzing their semantic connections with human poses.
SIF3D adeptly identifies salient points and the underlying ones in the 3D scene (as in Figure \ref{f1}), 
where the salient points are more likely to interact with the human, and thus the generated motion sequence is more realistic.
Additionally, inspired by \cite{wang2021scene}, we apply a geometry discriminator to boost the overall authenticity of the predicted motions.

Our contributions are as follows:
(1) We introduce SIF3D which considers both the external scene and the internal human gaze, and is capable of generating the accurate future motion, as well as the trajectory, within given 3D scenarios.
(2) Both semantic coherence-aware attention and ternary intention-aware attention are proposed, to explicitly distinguish salient point clouds to the local human pose representation, and the one to the global trajectory planning.
(3) We show that, on two benchmarks GIMO and GTA-1M, the proposed SIF3D achieves state-of-the-art performance both in human pose and trajectory prediction.

\vspace{-0.0em}
\section{Related Work}
\vspace{-0.0em}
\label{sec:2}
\tb{3D human motion prediction} has witnessed substantial advancements, 
particularly with the rise of deep end-to-end approaches \cite{Jain2016StructuralRNNDL,barsoum2018hp,kundu2018bihmp,Ruiz2018HumanMP,Guo2019HumanMP}.
Prior works typically leverage RNNs \cite{martinez2017human,martinez2021pose,fragkiadaki2015recurrent,Gui2018AdversarialGH,Gui2018TeachingRT,Tang2018LongTermHM}
to capture the motion temporal dependence, facing the limitations of discontinuity and error accumulation.
To address it, recent alternatives, CNN-based \cite{li2018convolutional} 
and transformer-based \cite{aksan2020attention} methods have been proposed.
To effectively exploit both spatial and temporal dependencies, \cite{aksan2020attention} designs a spatial-temporal attention schema, while \cite{xu2023auxiliary} proposes a dependencies modeling AuxFormer.
Notably, GCNs-based predictors are the dominant techniques and still evolve nowadays \cite{mao2019learning,cui2020learning,Cui_2021_CVPR,dang2021msr,li2020multitask,li2020dynamic,li2022skeleton}.

We also notice that current methods often fail to explicitly consider real 3D scenes, leading to physically implausible and artifactual predicted trajectories and poses. 
To solve these limitations, our SIF3D is proposed.

\tb{Scene-aware motion generation.} 
Considering the connection between human activities and scene context,
scene-aware motion generation has become a focal point of research 
\cite{corona2020context, cao2020long, hassan2021stochastic, zheng2022gimo}.
Early attempts \cite{corona2020context, cao2020long, wang2021scene} 
incorporate object bounding boxes, 2D scene images, or depth maps to contextualize scenes.
\cite{hassan2021stochastic} further introduces the object's geometry to estimate the interaction and generate interactive motions,
while adaptations of the A$*$ algorithm have been proposed for collision-averse trajectory planning \cite{hassan2021stochastic, wang2022towards}.

The adoption of directly accessible 3D point clouds for scene representation marks a new trend in scene-aware motion generation \cite{zheng2022gimo, wang2022humanise, huang2023diffusion, zhong2023rspt}.
\cite{wang2022humanise} proposes to condition the motion synthesis through text descriptions,
while \cite{huang2023diffusion} contributes a conditional generative diffusion model for scene understanding,
and \cite{zhong2023rspt} advances a structure-aware motion framework for active object tracking.
Notably, BiFu \cite{zheng2022gimo} introduces a bidirectional fusion strategy for motion prediction.
However, its neglect of local scene intricacies limits salient point detection, leading to ambiguity in cross-modal motion encoding across the scene.
In contrast, our SIF3D adeptly isolates critical scene points, enhancing the integration of pivotal scene data into motion forecasting.

\begin{figure*}[th!]
    \centering
    \vspace{-0.0em}
    \includegraphics[width=0.95\textwidth]{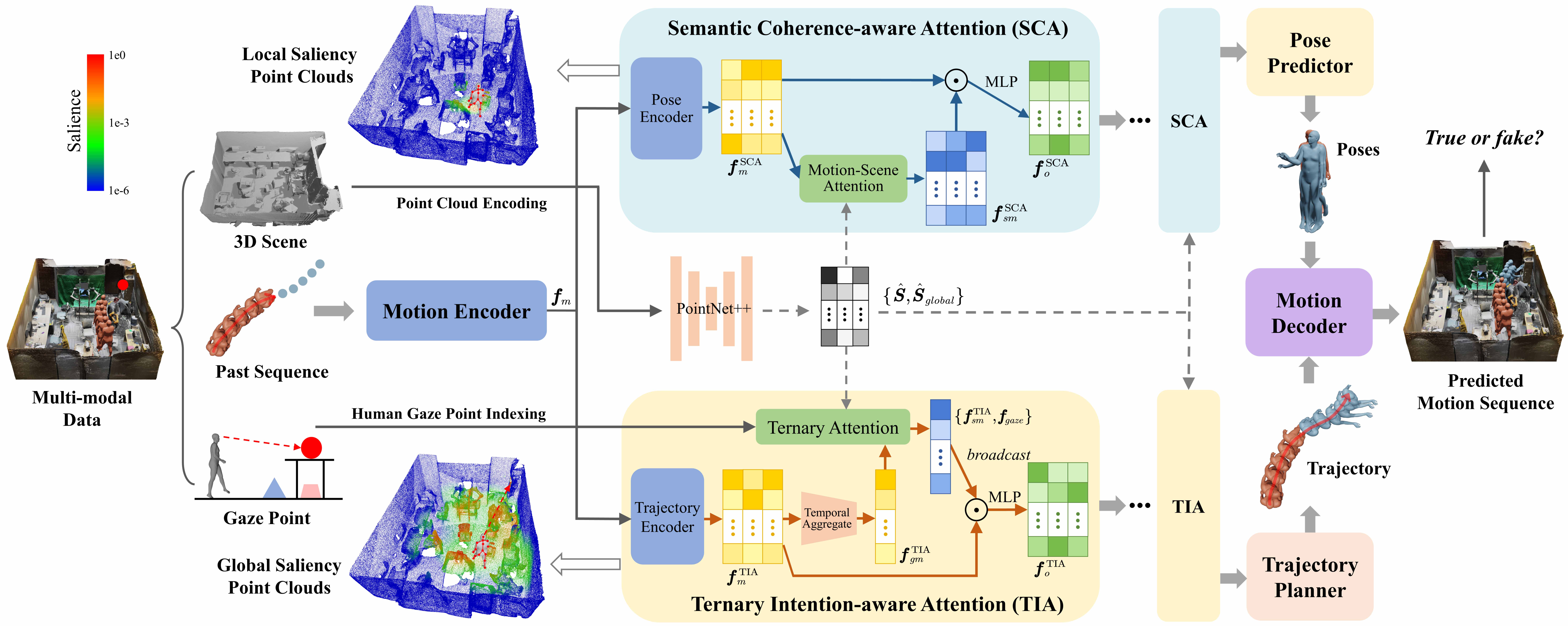}
    \vspace{-0.6em}
    \caption{
        \tb{The architecture of SIF3D.}
        SIF3D incorporates three modalities of input, the past motion sequence, the 3D scene point cloud, and the human gaze.
        First, MotionEncoder encodes past motion sequence into a motion embedding $\bs{f}_m$, 
        and the 3D scene $\bs{S}$ is encoded into $\{\bs{\hat{S}}, \bs{\hat{S}}_{global}\}$ through PointNet++ \cite{qi2017pointnet++}.
        Then, our TIA mechanism compresses motion embedding in the temporal dimension and searches for global salient points in the scene
        for trajectory planning. In addition, human gaze point $\bs{G}$ is introduced to index the scene point cloud for gaze point scene feature extraction.
        The SCA mechanism, on the other hand, is designed to capture local salient points in the scene for each independent pose.
        A TrajectoryPlanner and a PosePredictor are applied to predict trajectory and poses, respectively.
        And finally, the predicted motion sequence is generated through a MotionDecoder, which is supervised by the geometric discriminator.
    }
    \vspace{-1.0em}
    \label{f2}
  \end{figure*}

\tb{3D scene understanding.}
3D Point cloud, a vital geometric data structure, 
has been traditionally transformed into regular 3D voxel grids or collections of images, because of its irregular representation 
\cite{maturana2015voxnet, qi2016volumetric, wu20153d, qi2016volumetric, su2015multi}. 
However, these methods are either computationally expensive or task-specific, making them hard to transfer into downstream tasks.
To address it, PointNet and PointNet++ \cite{qi2017pointnet, qi2017pointnet++} introduces a unified architecture directly consuming point clouds, 
proving highly efficient and effective. 
Since then, various methods have been proposed for different tasks, \eg,
3D object detection \cite{gwak2020generative, qi2020imvotenet, qi2019deep}, 
semantic segmentation \cite{choy20194d, graham20183d, thomas2019kpconv}, 
and instance segmentation \cite{engelmann20203d, hou20193d, hou2020revealnet, jiang2020end}.

Although 3D scenes are crucial for information, effectively utilizing them in human motion prediction presents a challenge.
Previous works \cite{qi2017pointnet, qi2017pointnet++} have shown that many point cloud points are often superfluous or irrelevant.
SIF3D advances this understanding by selectively focusing on salient points within the 3D scene, leading to enhanced prediction quality for human poses and global trajectories.

\vspace{-0.0em}
\section{Proposed Method}
\vspace{-0.0em}
\label{sec:3}
Figure \ref{f2} illustrates the overall architecture of our SIF3D.
It begins with the proposed MotionEncoder and the vanilla PointNet++ \cite{qi2017pointnet++}, to encode the observed sequence, and 3D point clouds. 
Then, ternary intention-aware attention (TIA) and semantic coherence-aware attention (SCA) are used to extract crossmodal features from salient scene points, 
followed by a TrajectoryPlanner and a PosePredictor to plan future trajectories and predict poses, respectively.
Ultimately, the predicted motion sequence is generated by a MotionDecoder, supervised by the geometric discriminator.

\vspace{-0.2em}
\subsection{Problem Setup}
\vspace{-0.3em}
\label{sec:3.1}
Given $T$ historical motion $\bs{X}_{1:T} = \{ \bs{x}_1, \bs{x}_2, \cdots, \bs{x}_T \}$ 
in the scene cloud $\bs{S} \in \rr^{n \times 3}$,
corresponding with the human gaze point sequence $\bs{G}_{1:T} = \{\bs{g}_1, \bs{g}_2, \cdots, \bs{g}_T \}$,
we aim to predict the future motion $\bs{Y}_{1:\Delta T}$.
Here, $n$ denotes the size of the scene point clouds, $\bs{g}_k \in \rr^{3}$ is the position of human gaze
(denoted as the intersection of the human eye ray and 3D scene).
Each pose is described as $\bs{x}_k = (\bs{t}_k, \bs{o}_k, \bs{p}_k)$ \cite{zheng2022gimo},
where $\bs{t}_k \in \rr^{3}$ denotes the global translation, 
$\bs{o}_k \in SO(3)$ denotes the orientation, 
and $\bs{p}_k \in \rr^{32}$ refers to the body pose embedding.
The SMPL-X body mesh and the joints position $\bs{j}_k \in \rr^{23 \times 3}$ can be then obtained via VPoser \cite{pavlakos2019expressive}.
Following \cite{zheng2022gimo}, we set $T\!=\!3\text{-}sec$ and $\Delta T\!=\!5\text{-}sec$
to achieve a more challenging long-term prediction compared with the previous tasks ($\Delta T=1\text{-}sec$)
\cite{mao2019learning, li2022skeleton}.

In summary, with the model parameter $\theta$, our goal is:
\vspace{-0.4em}
\begin{equation}
    \label{eq:problem}
    \arg \max P(\bs{Y}_{1:\Delta T} |\bs{X}_{1:T}, \bs{S}, \bs{G}_{1:T}; \theta).
    \vspace{-0.4em}
\end{equation}

Note that the motion trajectory involves both global translation $\bs{T}_{1:T+\Delta T}$ and orientation $\bs{O}_{1:T+\Delta T}$.
Our primary focus in trajectory evaluation centers on $\bs{T}_{1:T+\Delta T}$ due to its significant impact on the overall motion. 
The orientation is implicitly reflected in the joints positions $\bs{J}_{1:T+\Delta T}$.

In contrast to numerous prior literature \cite{taheri2020grab, ionescu2013human3},
our task incorporates 3D scene and human gaze information, 
with the objective of producing semantically reasonable, and physically stable motion predictions within a given 3D scene.
Our effort aims to bridge the divide between motion prediction and real-world scenarios, fostering diverse applications grounded in realistic 3D scenes.

\subsection{Multimodal Encoding}
\vspace{-0.3em}
\label{sec:3.2}
In addition to the observed motion sequence $\bs{X}_{1:T}$, SIF3D integrates two additional input modalities: the external 3D scene $\bs{S}$, and the internal human's gaze $\bs{G}_{1:T}$.
The motion and the scene are first encoded into a unified space, while the gaze point is used for scene point indexing
to obtain the gaze point feature (see Sec-\ref{sec:3.3}).

A transformer-based MotionEncoder \cite{vaswani2017attention} is proposed to transfer the passing motion sequence $\bs{X}_{1:T}$ into a motion embedding $\bs{f}_m$.
Note that, instead of using the original observation $\bs{X}_{1:T}$ with $T$ frames, we pad it with the last observed pose $\bs{x}_{T}$ repeated $\Delta T$ times to construct the virtual sequence $\{\bs{X}, \bs{x}_T, ... , \bs{x}_T\} \in \rr^{(T+\Delta T) \times c_m}$, which is taken as the input of the MotionEncoder:
\vspace{-.3em}
\begin{equation}
    \label{eq:method_pre1}
    \bs{f}_m = \text{MotionEncoder}(\{\bs{X}, \bs{x}_T, ... , \bs{x}_T\}),
    \vspace{-0.4em}
\end{equation}
where the padded frames need to be predicted, and will be replaced after the inference to achieve the final prediction.
Here, $\bs{f}_m \in \rr^{(T+\Delta T)\times c_m}$, and $c_m$ is the dimension.

Meanwhile, 3D point cloud $\bs{S}$ is encoded using PointNet++ \cite{qi2017pointnet++}
to obtain per-point scene features $\bs{\hat{S}}$ and the global scene embedding $\bs{\hat{S}}_{global}$:

\vspace{-1.5em}
\begin{gather}
    \label{eq:method_pre2}
    \bs{\hat{S}}, \bs{\hat{S}}_{global} = \text{PointNet++}(\bs{S}),
\end{gather}
\vspace{-1.5em}

\noindent where $\bs{\hat{S}} \in \rr^{n \times c_s}$, $\bs{\hat{S}}_{global} \in \rr^{c_s}$,

\subsection{Ternary Intention-Aware Attention}
\vspace{-0.3em}
\label{sec:3.3}
Scene information serves as a crucial constrain in predicting the long-term trajectory 
\cite{corona2020context, cao2020long, wang2021scene}.
Moreover, as people often direct their gaze toward the objectives of their movements, 
human gaze becomes an essential cue, reflecting the potential destination of the motion trajectory \cite{zheng2022gimo}.

To enhance trajectory planning by effectively utilizing scene and gaze information, 
we propose ternary intention-aware attention (TIA).
TIA is designed to capture global salient scene features and analyze human intention through 3 distinct modalities: 
motion embedding $\bs{f}_m$, 3D scene $\{\bs{\hat{S}}, \bs{\hat{S}}_{global}\}$, and human gaze $\bs{G}_{1:T}$.

Since long-term trajectory planning hinges on the semantics of the entire motion sequence,
TIA first extracts a comprehensive global motion representation.
It encodes the input motion embedding to $\bs{f}^{\text{TIA}}_m$,
and then condenses it into a global motion representation $\bs{f}^{\text{TIA}}_{gm}$ through a temporal aggregator.
Various methods can be employed to design this aggregator, while we simply adopt the last motion embedding 
as the global representation in our case, aligning with a widely-used technique in NLP \cite{radford2021learning}:

\vspace{-1.5em}
\begin{align}
    \label{eq:tiaa1}
    \bs{f}^{\text{TIA}}_m &= \text{TrajectoryEncoder}(\bs{f}_m), \\
    \bs{f}^{\text{TIA}}_{gm} &= \text{Aggregate}(\bs{f}^{\text{TIA}}_m).
\end{align}
\vspace{-1.5em}

The global salient scene features are grounded in the global scene embedding $\bs{\hat{S}}_{global}$ as a foundation.
To achieve the extraction of effective global salient scene features,
TIA assesses the global salience $s_g \in \rr^{n}$ for each scene point to $\bs{f}^{\text{TIA}}_{gm}$, 
focusing only on the salient points.
To exclude underlying points, a simple method arises
by multiplying the salience with the corresponding scene point features.
This process yields the global salience $s_g \in \rr^{n}$ and the crossmodal global scene-motion features 
$\bs{f}^{\text{TIA}}_{sm}$ through a crossmodal attention mechanism as follows:

\vspace{-1em}
\begin{align}
    \label{eq:tiaa2}
    {\bs{Q}^{\text{TIA}}; \,  (\bs{K}^{\text{TIA}}, \bs{V}^{\text{TIA}}) = \text{Linear}(\bs{f}^{\text{TIA}}_{gm}); \, \text{Linear}(\bs{\hat{S}})}, \\
    s_g = \text{softmax}\bigg(\dfrac{\bs{Q}^{\text{TIA}} * (\bs{K}^{\text{TIA}})^T}{\sqrt{c}}\bigg), \\
    \bs{f}^{\text{TIA}}_{sm} = \text{broadcast}(\bs{\hat{S}}_{global} + s_g \cdot \bs{V}^{\text{TIA}}),
\end{align}
\vspace{-1.5em}

\noindent where $c$ is the attention dimension. 
We broadcast the global salient scene features to the same shape as $\bs{f}_m$ 
to disperse it across the entire motion sequence.

Moreover, recognizing the informative nature of the gaze regarding the subject's intent, 
we also introduce the human gaze into cross-modal attention.
The gaze point $\bs{G}_{1:T}$, defined as the intersection between the human eye's line of sight and the 3D scene, 
correlates with a specific scene point.
By indexing the scene point cloud features $\bs{\hat{S}}$ with the gaze point, 
we derive the gaze-related scene feature, representing the potential human intention. 
This $T$-frame gaze feature (padded with last frame for $\Delta T$ times to fit the length) 
is then encoded to align with the length of motion sequence using a GazeEncoder:

\vspace{-1.5em}
\begin{gather}
    \label{eq:tiaa3}
    \bs{f}_{gaze} = \text{GazeEncoder}(\{\bs{\hat{S}}[\bs{G}], \bs{\hat{S}}[\bs{g}_T], ... , \bs{\hat{S}}[\bs{g}_T]\}), 
\end{gather}
\vspace{-1.5em}

\noindent where $[\cdot]$ represents the indexing operation, $\bs{f}_{gaze}$ represents the encoded gaze features.
Finally, TIA integrates the features from the three modalities: $\bs{f}_{m}$ for motion, 
$\bs{f}^{\text{TIA}}_{sm}$ for scene, and $\bs{f}_{gaze}$ for gaze.
The output of TIA is then computed through a standard two-layer MLP \cite{vaswani2017attention}:

\vspace{-1.5em}
\begin{gather}
    \label{eq:tiaa4}
    \bs{f}^{\text{TIA}}_o = \text{MLP}(\text{concat}(\bs{f}_m, \bs{f}^{\text{TIA}}_{sm}, \bs{f}_{gaze})).
\end{gather}
\vspace{-1.5em}

\vspace{-0.4em}
\subsection{Semantic Coherence-Aware Attention}
\vspace{-0.3em}
\label{sec:3.4}
We introduce semantic coherence-aware attention (SCA) to integrate local salient scene details into per-frame pose prediction.
In contrast to TIA, SCA computes scene point salience for each frame independently.
It initiates with a PoseEncoder to encode the input motion embedding $\bs{f}_m$ into pose embedding $\bs{f}^{\text{SCA}}_m$. 
Then, the local scene point salience $s_l \in \rr^{(T+\Delta T) \times n}$ is evaluated between the scene features $\bs{\hat{S}}$
and each of the $T+\Delta T$ pose embedding:

\vspace{-1.5em}
\begin{gather}
    \label{eq:scaa1}
    \bs{f}^{\text{SCA}}_m = \text{PoseEncoder}(\bs{f}_m), \\
    {\bs{Q}^{\text{SCA}}; \, \bs{K}^{\text{SCA}}, \bs{V}^{\text{SCA}} = \text{Linear}(\bs{f}^{\text{SCA}}_{m}), \, \text{Linear}(\bs{\hat{S}})}, \\
    s_l = \text{softmax}\bigg(\dfrac{\bs{Q}^{\text{SCA}} * (\bs{K}^{\text{SCA}})^T}{\sqrt{c}}\bigg),
\end{gather}
\vspace{-1.5em}

\noindent where $c~$ represents the attention dimension. 

SCA places a greater emphasis on fine-grained local scene information, exhibiting heightened sensitivity to spatial details.
In practice, points situated closer to the subject and aligned with the human body are empirically considered more salient.
To adopt this spatial attention mechanism, 
we incorporate a spatial salience bias to $s_l$, denoted as $s_{spatial} \in \rr^{(T+\Delta T) \times n}$,
inspired by \cite{liu2021swin, liu2022swin, dong2022cswin}.
The absolute scene points position $\bs{S}$ is normalized to the relative positions 
$\bs{S}_{rel} \in \rr^{(T+\Delta T) \times n \times 3}$ 
according to the predicted human translation $\bs{\hat{T}}$ and orientation $\bs{\hat{O}}$ from TrajectoryPlanner.
Then, the spatial salience bias $s_{spatial}~$ is determined based on $\bs{S}_{rel}$ through a standard two-layer MLP:

\vspace{-1.5em}
\begin{gather}
    \label{eq:scaa2}
    \bs{S}_{rel} = \text{Normalize}(\bs{S}; \bs{\hat{T}}, \bs{\hat{O}}), \\
    s_{spatial} = \text{MLP}(\bs{S}_{rel}).
\end{gather}
\vspace{-1.5em}

\noindent The local salient scene features are then extracted based on locally salient scene points, 
while the low-salience underlying points are ignored through salience multiplication:

\vspace{-1.5em}
\begin{gather}
    \label{eq:scaa3}
    \bs{f}^{\text{SCA}}_{sm} = (s_l + s_{spatial}) \cdot \bs{V}^{\text{SCA}}.
\end{gather}
\vspace{-1.5em}

\noindent Finally, SCA produces output by combining the motion features $\bs{f}_m$ 
with the local salient scene features $\bs{f}^{\text{SCA}}_{sm}$:

\vspace{-1.5em}
\begin{gather}
    \label{eq:scaa4}
    \bs{f}^{\text{SCA}}_o = \text{MLP}(\text{concat}(\bs{f}^{\text{SCA}}_m, \bs{f}^{\text{SCA}}_{sm})).
\end{gather}
\vspace{-1.5em}

\vspace{-0.2em}
\subsection{Motion Sequence Generation}
\vspace{-0.2em}
\label{sec:3.5}
Predicting future motion involves considering both the trajectory and poses.
In our way of multimodal feature extraction, TIA adds global scene features and human intention, 
while SCA brings in local scene details.
Since trajectory planning looks further ahead, and human poses are more related to the local scene semantics, 
it makes sense in practice that TIA could enhance trajectory prediction, while SCA helps predict poses.
Thus we use the features from TIA to predict the global translation $\bs{\hat{T}}$
and orientation $\bs{\hat{O}}$, while relying on SCA for prediction poses $\bs{\hat{P}}$:

\vspace{-1.5em}
\begin{gather}
    \label{eq:mf0}
    \bs{\hat{T}}, \bs{\hat{O}}=\text{TrajectoryPlanner}(\bs{f}^{\text{TIA}}_o), \\
    \bs{\hat{P}}=\text{PosePredictor}(\bs{f}^{\text{SCA}}_o),
\end{gather}
\vspace{-1.5em}

\noindent where TrajectoryPlanner/PosePredictor are linear layers.

The predicted trajectory and pose are then combined to generate the final motion sequence $\bs{\hat{Y}}$ through a motion decoder.
To achieve precise adjustment of joint positions, 
the motion decoder applies a GCN architecture \cite{kipf2016semi}, 
and we calculate the SMPL-X human skeleton based on the predicted trajectory and pose embedding
through the frozen pre-trained SMPL-X model and VPoser:

\begin{table*}[t!]
    \renewcommand\arraystretch{0.9}
    \centering
    \footnotesize
    \setlength{\tabcolsep}{1.5mm}{
    \begin{tabular}{c|cccc|cccc}
    \multirow{2}{*}{Dataset} &
    \multicolumn{4}{c|}{GIMO \cite{zheng2022gimo}} & \multicolumn{4}{c}{GTA-1M \cite{cao2020long}}  \\
    &Traj-path &Traj-dest &MPJPE-path &MPJPE-dest &Traj-path &Traj-dest &MPJPE-path &MPJPE-dest \\
    \hline
    {LTD~\cite{mao2019learning}} 
    &681 &890 &158.7 &204.6
    &691 &1051 &175.0 &251.8
    \\
    {SPGSN~\cite{li2022skeleton}} 
    &739 &910 &159.5 &203.1
    &805 &1115 &167.3 &241.4
    \\
    {AuxFormer~\cite{xu2023auxiliary}}
    &688 &893 &173.5 &205.4
    &688 &1078 &187.8 &262.0
    \\
    \tb{SIF3D~} 
    &737 &888 &169.6 &205.6
    &772 &1072 &182.0 &250.2
    \\
    \hline
    {LTD~\cite{mao2019learning}}+Scene 
    &673 &890 &156.4 &203.0
    &690 &1029 &175.8 &245.6
    \\
    {SPGSN~\cite{li2022skeleton}}+Scene 
    &700 &893 &157.0 &202.2
    &767 &1054 &167.8 &236.9
    \\
    {AuxFormer~\cite{xu2023auxiliary}}+Scene 
    &732 &902 &165.1 &202.4
    &745 &1028 &167.0 &246.0
    \\
    \tb{SIF3D~~w/ Scene}
    &700 &823 &164.2 &199.6
    &746 &967 &175.2 &231.4
    \\
    \hline
    {LTD~\cite{mao2019learning}}+Gaze 
    &655 &807 &157.6 &203.5
    &687 &996 &169.1 &246.0
    \\
    {SPGSN~\cite{li2022skeleton}}+Gaze 
    &681 &857 &158.2 &201.7
    &748 &1031 &164.4 &234.8
    \\
    {AuxFormer~\cite{xu2023auxiliary}}+Gaze
    &627 &755 &165.8 &202.8
    &672 &932 &176.0 &256.6
    \\
    \tb{SIF3D~w/ Gaze}
    &603 &714 &163.1 &199.1
    &680 &917 &171.5 &233.0
    \\
    \hline
    {LTD~\cite{mao2019learning}}+Scene+Gaze
    &655 &801 &\tb{155.6} &202.0
    &702 &982 &167.8 &241.9
    \\
    {SPGSN~\cite{li2022skeleton}}+Scene+Gaze 
    &702 &982 &156.9 &202.0
    &737 &1018 &\tb{162.7} &238.1
    \\
    {AuxFormer~\cite{xu2023auxiliary}}+Scene+Gaze
    &622 &746 &161.5 &201.1
    &672 &916 &165.3 &236.8
    \\
    {BiFu~\cite{zheng2022gimo}}~w/ Scene+Gaze
    &661 &727 &167.8 &205.0
    &683 &903 &164.6 &234.2
    \\
    \tb{SIF3D~w/ Scene+Gaze}
    &\tb{590} &\tb{666} &156.6 &\tb{195.7}
    &\tb{626} &\tb{836} &164.9 &\tb{227.7}
    \\
    \hline
    \end{tabular}
    }
    \vspace{-0.7em}
    \caption{
    \tb{Comparison of trajectory deviation and MPJPE} (in $mm$) over the sequences of the GIMO \cite{zheng2022gimo} 
    and GTA-1M \cite{cao2020long} datasets.
    The best result is highlighted in bold.
    From the results, we observe that the 3D scene and gaze information can boost all the methods,
    and the proposed SIF3D obtains both lower trajectory deviation and smaller MPJPE
    in almost all scenarios compared to the previous methods.
}
\label{t1}
\vspace{-0.0em}
\end{table*}

\vspace{-1.5em}
\begin{gather}
    \label{eq:mf1}
    \bs{\hat{J}}=\text{SMPL-X}(\bs{\hat{T}}, \bs{\hat{O}}, \bs{\hat{P}}, \text{VPoser}(\bs{\hat{P}})), \\
    \bs{\hat{Y}}=\text{MotionDecoder}(\bs{\hat{J}}),
\end{gather}
\vspace{-1.5em}

\noindent where $\bs{\hat{J}}$ represents the reconstructed joints positions,
and $\bs{\hat{Y}}$ denotes the final predicted motion sequence.

To further improve the physical stability of the predicted motion sequence,
a geometric discriminator is introduced following \cite{wang2021scene} to classify the fake and true motion sequences to the ground truth within the 3D scenarios.

\vspace{-0.2em}
\subsection{Implement Details}
\vspace{-0.2em}
\label{sec:3.6}
\tb{Architecture Details:}
In our experiments, we employ a 6-layer transformer encoder for the MotionEncoder,
a single-layer self-attention mechanism for both the TrajectoryEncoder and the PoseEncoder,
and a 6-layer graph convolutional network for the MotionDecoder.
We configured the stack size of both TIA and SCA to be 2, indicating that SIF3D incorporates 2 TIA-blocks and 2 SCA-blocks.
Furthermore, we set all embedding dimensions, including attention dimension $c$, motion embedding dimension $c_m$,
and scene embedding dimension $c_s$, to 256.
The MLPs in TIA and SCA are both 2-layer feed-forward with a hidden dimension of 1024.
Finally, the geometry discriminator is designed with a 3-layer transformer decoder.

\tb{Training Setup:}
We train our model using an AdamW optimizer with an initial learning rate of 0.0004.
An exponential learning rate scheduler is applied with a decay rate of 0.98.
The models are all trained for 100 epochs with a batch size of 8 on a single NVIDIA RTX3090 GPU.

\vspace{-0.3em}
\section{Experiments}
\vspace{-0.2em}
\label{sec:4}

\vspace{-0.0em}
\subsection{Experimental Setup}
\label{sec:4.1}
\vspace{-0.3em}

\tb{Dataset-1:  GIMO} \cite{zheng2022gimo} records three modalities, including (1) full-body SMPL-X poses with $\approx$ 129$K$ frames.
(2) 3D geometry scene scanned from LiDAR sensors.
(3) human gaze is recorded as a 3D coordinate.
We down-sample the motion sequences to 2fps for long-term prediction and maintain the original train-test split \cite{zheng2022gimo}.

\tb{Dataset-2: GTA-1M} \cite{cao2020long} comprises over 1000$K$ frames, which collects the human pose from real game engine rendering. 
The dataset has clean 3D human pose and camera annotations, and large diversity in human appearances, camera views: 10 large houses, 13 weathers, 50 human models, 22 walking styles, and various actions.

\tb{Baselines:}
Our SIF3D is compared with 4 recent methods, \ie, LTD \cite{mao2019learning}, SPGSN \cite{li2022skeleton}, BiFu \cite{zheng2022gimo}
and AuxFormer \cite{xu2023auxiliary}.
BiFu \cite{zheng2022gimo} and AuxFormer\cite{xu2023auxiliary} are transformer-based, while LTD \cite{mao2019learning} and SPGSN\cite{li2022skeleton} are variants of GCNs.
For a fair comparison, for the methods that do not inherently consider 3D scenes or human gaze, we concatenate the global scene embedding $\bs{\hat{S}}_{global}$ to the motion sequence and include the gaze embedding $\bs{f}_{gaze}$. 
Other aspects align with their open-source codes and original settings.

\tb{Metrics:}
We evaluate SIF3D from 2 main aspects: trajectory and human pose.
The trajectory evaluation \cite{zheng2022gimo,altche2017lstm,nikhil2018convolutional} 
quantifies the difference between the predicted trajectory and the ground truth one. 
The MPJPE \cite{mao2019learning, li2022skeleton,aksan2021spatio,zheng2022gimo}
evaluates the mean error in the position across all human joints. 
Specifically, the result is analyzed from the intermediate poses and the final destination:
\tb{(1)~Traj-path}: the average trajectory deviation over all predicted poses;
\tb{(2)~Traj-dest}: the trajectory deviation of the end pose;
\tb{(3)~MPJPE-path}: the average MPJPE over the predicted sequence;
\tb{(4)~MPJPE-dest}: the MPJPE of the destination.

\begin{table*}[t!]
    \vspace{-0em}
    \renewcommand\arraystretch{1.1}%行距
    \centering
    \scriptsize
    \setlength{\tabcolsep}{0.68mm}{
    \begin{tabular}{c|c|c|ccc|ccc|ccc|ccc|ccc|ccc|ccc}
    &\multirow{2}{*}{Method} & Gaze\& & \multicolumn{3}{c|}{bedroom} &\multicolumn{3}{c|}{classroom} &\multicolumn{3}{c|}{garden} 
    &\multicolumn{3}{c|}{kitchen} &\multicolumn{3}{c|}{lab} &\multicolumn{3}{c|}{livingroom} &\multicolumn{3}{c}{seminarroom}\\
    \cline{4-24}
    &&Scene &0.5$s$ &2.0$s$ &5.0$s$ &0.5$s$ &2.0$s$ &5.0$s$ &0.5$s$ &2.0$s$ &5.0$s$ &0.5$s$ &2.0$s$ &5.0$s$ &0.5$s$ &2.0$s$ &5.0$s$ &0.5$s$ &2.0$s$ &5.0$s$ &0.5$s$ &2.0$s$ &5.0$s$ \\
    \hline
    \multirow{8}{*}{\rotatebox{90}{Traj}}
    &\multirow{2}{*}{LTD~\cite{mao2019learning}}
    &$\times$ &109 &459 &1147 &195 &415 &613 &149 &578 &1076 &100 &457 &481 &184 &644 &1165 &\tb{127} &624 &1212 &190 &529 &851 \\
    &&\checkmark &\tb{108} &\tb{411} &966 &192 &422 &524 &147 &496 &560 &\tb{88} &528 &589 &149 &516 &878 &133 &\tb{551} &1297 &\tb{170} &527 &750 \\
    \cline{2-24}
    &\multirow{2}{*}{SPGSN~\cite{li2022skeleton}} 
    &$\times$ &144 &492 &922 &189 &\tb{396} &564 &\tb{129} &556 &813 &117 &639 &660 &181 &640 &1014 &153 &703 &1454 &182 &589 &866 \\
    &&\checkmark &112 &475 &965 &180 &407 &497 &172 &540 &661 &156 &640 &677 &\tb{120} &515 &919 &135 &649 &1258 &174 &514 &753 \\
    \cline{2-24}
    &\multirow{2}{*}{AuxFormer~\cite{xu2023auxiliary}} 
    &$\times$ &185 &541 &1162 &319 &442 &529 &232 &630 &972 &172 &509 &625 &248 &745 &1113 &151 &621 &1030 &273 &606 &881 \\
    &&\checkmark &212 &472 &813 &228 &427 &532 &231 &536 &577 &225 &544 &555 &246 &\tb{511} &\tb{621} &273 &647 &1082 &222 &546 &735 \\
    \cline{2-24}
    &{BiFu~\cite{zheng2022gimo}}
    &\checkmark &232 &540 &838 &332 &442 &689 &218 &532 &659 &275 &\tb{467} &582 &273 &551 &681 &340 &607 &936 &417 &659 &629 \\
    \cline{2-24}
    &\tb{SIF3D}
    &\checkmark &183 &441 &\tb{554} &\tb{155} &424 &\tb{431} &200 &\tb{472} &\tb{533} &176 &471 &\tb{444} &245 &540 &730 &257 &602 &\tb{794} &219 &\tb{440} &\tb{611} \\
    \hline
    \hline
    \multirow{8}{*}{\rotatebox{90}{MPJPE}}
    &\multirow{2}{*}{LTD~\cite{mao2019learning}}
    &$\times$ &92.9 &142.8 &263.5 &98.0 &160.4 &169.7 &73.6 &231.1 &209.1 &86.6 &102.3 &141.4 &70.3 &99.1 &222.6 &97.6 &128.6 &252.5 &109.5 &150.8 &204.2 \\    
    &&\checkmark &85.5 &134.9 &236.6 &110.1 &192.8 &159.8 &67.8 &227.0 &204.6 &91.0 &97.7 &128.3 &74.3 &104.1 &237.1 &111.3 &140.7 &271.7 &117.2 &156.0 &199.9 \\    
    \cline{2-24}
    &\multirow{2}{*}{SPGSN~\cite{li2022skeleton}} 
    &$\times$ &89.0 &131.4 &232.8 &102.9 &170.5 &160.5 &\tb{62.3} &218.8 &205.5 &87.5 &97.1 &131.6 &66.8 &98.5 &240.1 &103.6 &124.1 &285.2 &110.5 &149.2 &191.5 \\
    &&\checkmark &89.4 &137.6 &249.5 &101.7 &176.1 &147.8 &67.1 &233.2 &202.6 &\tb{85.5} &\tb{92.1} &120.9 &71.8 &98.7 &238.2 &109.1 &124.9 &281.3 &112.8 &145.4 &193.1 \\
    \cline{2-24}
    &\multirow{2}{*}{AuxFormer~\cite{xu2023auxiliary}} 
    &$\times$ &114.9 &173.2 &259.7 &137.1 &178.7 &162.4 &105.3 &230.5 &229.9 &128.5 &108.0 &125.1 &113.1 &144.0 &231.9 &156.3 &132.7 &259.5 &130.9 &160.7 &204.9 \\
    &&\checkmark &90.5 &142.3 &240.7 &110.2 &183.8 &170.5 &91.5 &232.7 &203.9 &103.8 &106.4 &115.5 &76.3 &112.9 &218.2 &113.5 &127.1 &271.5 &105.7 &146.0 &191.6 \\
    \cline{2-24}
    &{BiFu~\cite{zheng2022gimo}}
    &\checkmark &\tb{67.0} &115.0 &258.0 &\tb{86.6} &\tb{144.2} &171.9 &89.4 &205.0 &251.7 &98.0 &104.0 &133.0 &53.9 &110.8 &229.8 &\tb{95.7} &126.1 &236.1 &107.2 &\tb{128.7} &209.9 \\    
    \cline{2-24}
    &\tb{SIF3D}
    &\checkmark &82.0 &\tb{105.7} &\tb{185.9} &100.3 &147.1 &\tb{160.1} &74.0 &\tb{190.4} &\tb{199.0} &100.1 &101.5 &\tb{114.3} &\tb{65.4} &\tb{94.3} &\tb{218.1} &108.9 &\tb{121.1} &\tb{188.4} &\tb{101.8} &137.3 &\tb{179.4} \\    
    \hline

    \end{tabular}
    }
    \vspace{-1em}
    \caption{
        \tb{Performance details on the GIMO test set} \cite{zheng2022gimo} using trajectory deviation and MPJPE (in $mm$).
        Our SIF3D demonstrates strong performance across all scenarios, particularly excelling in long-term predictions, and achieves the best result on average.
    }
    \label{t2}
    \vspace{-1.5em}
  \end{table*}

\vspace{-0.em}
\subsection{Ablation of Multiple Modalities}
\vspace{-0.2em}
\label{sec:4.2}

The results in Table \ref{t1} highlight the vital role of gaze and 3D scene information in scene-conditioned motion prediction.
Note that, as GTA-1M lacks human gaze information, we approximate it using the intersection point of the ray facing the human face and the 3D scene.
We observe that the introduction of gaze and scene information improves the performance of all methods, 
including the estimated gaze information in GTA-1M, with SIF3D benefiting the most.

SIF3D excels by adeptly utilizing 3D scene information, surpassing baseline methods that show only marginal gains.
This success is attributed to SIF3D's proficient integration of scene features from both global and local perspectives 
through TIA and SCA.
On the other hand, the human gaze significantly enhances trajectory planning by providing key insights into the subject's intentions 
and closely aligning with the motion sequence's destination, resulting in a notable decrease in trajectory deviation.

While the introduction of both scene and gaze enhances the performance individually, 
their simultaneous incorporation yields optimal results across the board. 
It evidences the effectiveness of collaborating external scene and internal gaze features, 
establishing a robust foundation for improved motion prediction.

\vspace{-0.0em}
\subsection{Detailed Results}
\vspace{-0.3em}
\label{sec:4.3}

\begin{figure*}[th!]
    \centering
    \vspace{-0em}
    \includegraphics[width=0.99\textwidth]{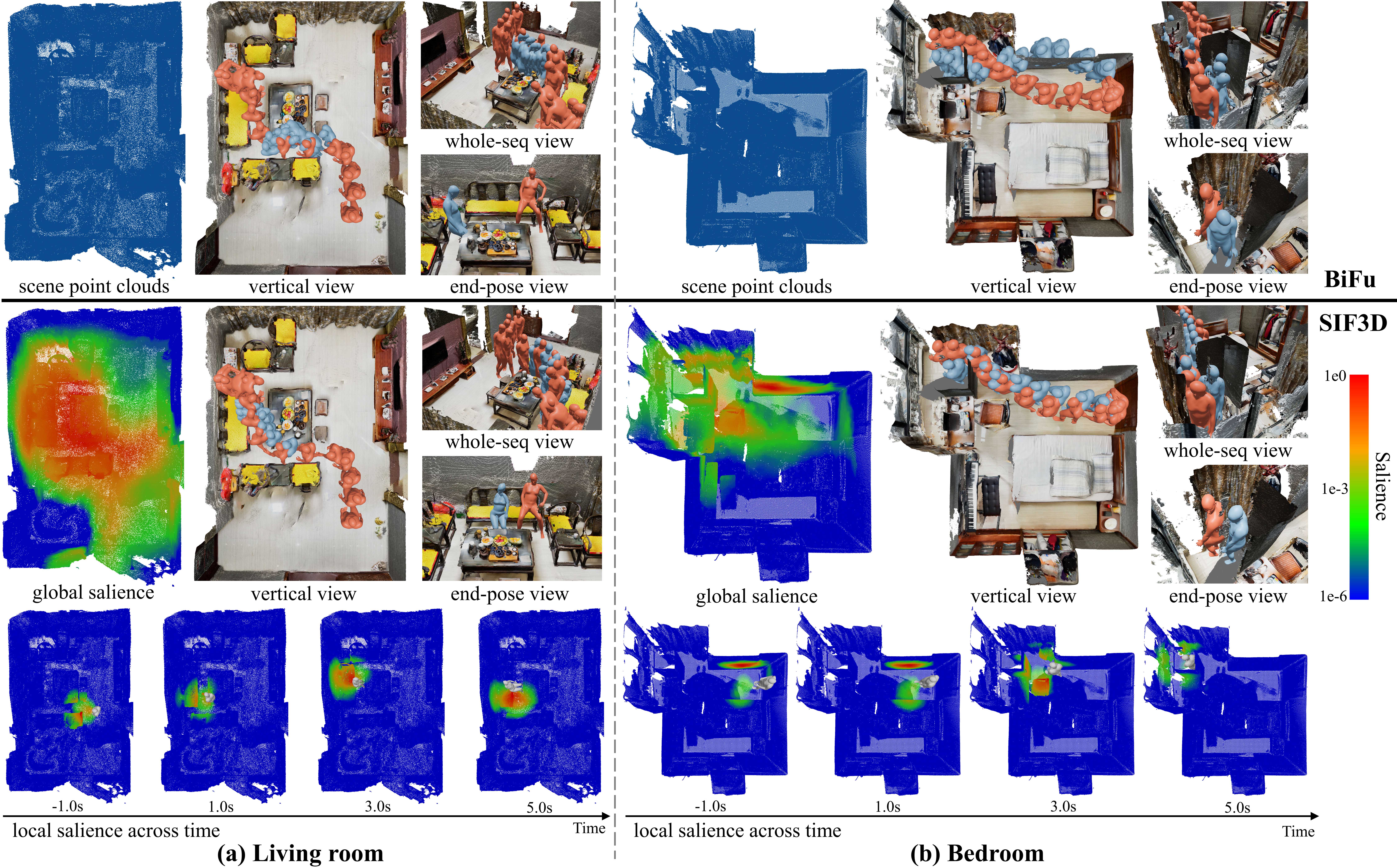}
    \vspace{-0.2em}
    \caption{
        \tb{Visualizations of our SIF3D compared with the SoTA BiFu,} under the scenarios of (a) living room and (b) bedroom.
        The top is the results of BiFu \cite{zheng2022gimo}, which equally treats all scene points; in contrast, the middle row is our SIF3D, where the salient points are highlighted in red, and the underlying points are in blue. 
        For the sake of clarity, the predicted sequence is presented from the vertical view, whole-seq view, and end-pose view.
        We note that the red human meshes are the ground truth, while the blue ones indicate the predictions.
        At the bottom, we present the local scene salience heatmap across time for SIF3D, with a time interval of 2 seconds.
    }
    \vspace{-1em}
    \label{f3}
  \end{figure*}

Next, we analyze the performance of each method under different timestamps, as illustrated in Table \ref{t2} and Table \ref{t2.5}.

From the result, we observe that AuxFormer \cite{xu2023auxiliary} demonstrates substantial improvement, especially in long-term trajectory predictions. 
BiFu \cite{zheng2022gimo} outperforms baseline methods lacking scene and gaze information and remains competitive with methods that include such information.
We attribute this success to the better ability of cross-modal feature extraction of transformers.
Similarly, our SIF3D also applies an attention mechanism (TIA and SCA), ensuring the effective extraction of multi-modal features.

GCN-based methods (LTD \cite{mao2019learning}, SPGSN \cite{li2022skeleton}) show proficiency in short-term predictions.
However, challenges arise in longer-term scenarios, particularly in 5-second predictions, where Traj-dest averages over 400mm for both LTD and SPGSN.
This difficulty arises due to the challenges GCN-based methods encounter in obtaining comprehensive whole-person pose features, 
making it challenging to estimate global human intention and establish effective cross-modal interaction with scene features.

\begin{table}[t!]
    \vspace{-0em}
    \renewcommand\arraystretch{0.96}
    \centering
    \footnotesize
    \setlength{\tabcolsep}{1.07mm}{
    \begin{tabular}{c|c|ccccc}
    &Method &0.5$s$ &1.0$s$ &2.0$s$ &3.0$s$ &5.0$s$ \\
    \hline
    \multirow{4}{*}{\rotatebox{90}{Traj}}
    &{AuxFormer~\cite{xu2023auxiliary}}
    &212 &322 &500 &771 &1078 \\
    &{AuxFormer~\cite{xu2023auxiliary}}+Scene+Gaze
    &198 &274 &411 &747 &916 \\
    \cline{2-7}
    &{BiFu~\cite{zheng2022gimo}}
    &192 &\tb{251} &417 &647 &903 \\
    \cline{2-7}
    &\tb{SIF3D}
    &\tb{179} &265 &\tb{392} &\tb{579} &\tb{836} \\
    \hline
    \multirow{4}{*}{\rotatebox{90}{MPJPE}}
    &{AuxFormer~\cite{xu2023auxiliary}}
    &105.3 &135.7 &163.7 &211.5 &262.0 \\
    &{AuxFormer~\cite{xu2023auxiliary}}+Scene+Gaze
    &\tb{95.5} &127.3 &160.7 &203.4 &236.8 \\
    \cline{2-7}
    &{BiFu~\cite{zheng2022gimo}}
    &100.3 &131.0 &158.8 &196.6 &234.2 \\    
    \cline{2-7}
    &\tb{SIF3D}
    &97.9 &\tb{125.7} &\tb{155.9} &\tb{192.4} &\tb{227.7} \\  
    \hline
    \end{tabular}
    }
    \vspace{-0.6em}
    \caption{
        \tb{Performance details on the GTA-1M test set} \cite{cao2020long} using trajectory deviation and MPJPE (in $mm$).
        We observe that our SIF3D consistently outperforms the baseline methods in motion prediction within game engine-rendered 3D scenarios.
    }
    \label{t2.5}
    \vspace{-1.8em}
    \end{table}

In contrast, our SIF3D achieves superior performance across all scenarios,
particularly achieving the best metrics in almost all long-term (5-sec) evaluations. 
This success is evident in both global trajectory and local pose, 
highlighting the effectiveness of TIA's intention understanding and SCA's locally scene-aware pose prediction. 
It also evidences the importance of scene-awareness and line of gaze-guidance for motion prediction in 3D scenes.

\vspace{-0.0em}
\subsection{Visualizations}
\vspace{-0.3em}
\label{sec:4.4}

To delve deeper into SIF3D, we provide the prediction visualizations in two different scenarios, where BiFu \cite{zheng2022gimo} is used as a baseline, due to its built-in integration of scene and gaze, resulting in superior numerical performance.

In the first scene (Figure \ref{f3}\tb{(a)}), the person navigates through the aisle between the chair and table, proceeding toward the sofa.
While BiFu estimates the overall movement trend, the predicted motion has the person passing through the coffee table. 
Moreover, in the final pose, the person appears suspended before the chair, deviating from physical constraints.
In contrast, SIF3D demonstrates exhibits precise predictions of the individual's motion.
The global scene point salience heatmap derived highlights SIF3D's ability to comprehend the 3D scene, 
concentrating on key interactive objects (chairs, sofa, and coffee table)
aligned with the observed motion sequence and corresponding human gaze.
This focus enables SIF3D to predict a clear human intention and devise a plausible trajectory plan.

In the second scene (Figure \ref{f3}\tb{(b)}), the individual is poised to traverse the hallway leading to the main entrance, reaching the upper left corner intending to open the window.
BiFu \cite{zheng2022gimo} predicts an accurate destination, but the motion sequence presents a significant physical inconsistency within the scene, penetrated by the wall (frames 2-5) and the screen (frames 7-8).
The global salience heatmap of SIF3D reveals the understanding of the semantics of the entire 3D scene, even if the upper left corner of the room is not visible at the end of the observation, providing a better action prediction.
In addition, the local salience heatmap also shows SIF3D's awareness of obstacles like the walls.

The visualizations demonstrate that SIF3D effectively predicts motions that align closely with the ground truth, 
accurately capturing both global trajectories and local poses.

\vspace{-0.0em}
\subsection{Ablation Studies}
\vspace{-0.0em}
\label{sec:4.5}

\tb{Ablation of each component.}
We first investigate the impact of key components within our SIF3D, including \tb{(1) TIA}, \tb{(2) SCA}, \tb{(3) MotionDecoder}, \tb{(4) Discriminator}, and \tb{(5) PointNet++}, as shown in Table \ref{t3}.

\begin{table}[h!]
    \vspace{-0.0em}
    \renewcommand\arraystretch{0.9}
    \centering
    \footnotesize
    \setlength{\tabcolsep}{1.0mm}{
    \begin{tabular}{c|cccc}
        &Traj-path &Traj-dest &MPJPE-path &MPJPE-dest \\
        \hline
        w/o TIA &701 &850 &167.9 &203.8 \\
        w/o SCA &640 &703 &164.8 &200.4 \\
        w/o MotionDecoder &669 &699 &159.2 &198.9 \\
        w/o Discriminator &604 &677 &158.2 &196.8 \\
        w/o PointNet++ &766 &918 &170.4 &218.5 \\
        \tb{SIF3D} &\tb{590} &\tb{666} &\tb{156.6} &\tb{195.7} \\
        \hline
    \end{tabular}
    }
    \vspace{-0.3em}
    \caption{
        \tb{SIF3D performance with ablations of components}, evaluated using trajectory deviation and MPJPE (in $mm$).
    }
    \label{t3}
  \vspace{-1.2em}
\end{table}

The removal of TIA and SCA results in a significant performance decline.
Specifically, \ti{w/o} TIA, SIF3D experiences a significant increase of 111mm in Traj-dest, underscoring a reduced capability to fully understand the 3D scene and human intentions.
Note that SCA's spatial salience bias $s_{spatial}$ depends on the predicted trajectory,
removing TIA also negatively impacts SCA due to the inaccurate predicted trajectory, 
which is evident in increases of 11.3mm and 8.1mm in MPJPE-path and MPJPE-dest.
In addition, \ti{w/o} SCA, SIF3D cannot capture local scene details, 
resulting in a 37mm increase in Traj-dest and a 4.7mm rise in MPJPE-dest.

TIA and SCA make independent predictions for global trajectory and local pose. 
Eliminating the motion decoder makes the predicted trajectory and pose identical,
which introduces a 20.0mm increase in Traj-dest and a 3.2mm rise in MPJPE-dest.
The discriminator also contributes to a 7.8mm decrease in Traj-dest and a 1.1mm decrease in MPJPE-dest.

Moreover, the scene encoder PointNet++, plays a crucial role as a vital link connecting our motion predictor. 
Notably, when PointNet++ is removed (replaced with an MLP), 
there is a remarkable increase in Traj-dest and MPJPE-dest. 
In summary, these results highlight the effectiveness of each component in our SIF3D.

\begin{table}[h!]
    \vspace{-0.2em}
    \renewcommand\arraystretch{1}
    \centering
    \footnotesize
    \setlength{\tabcolsep}{1.0mm}{
    \begin{tabular}{c|ccccccc}
        Point cloud size &512 &1024 &2048 &\tb{4096} &8192 &16384 &32768 \\
        \hline
        Traj-path~(mm) &626 &605 &594 &{590} &588 &587 &587 \\
        MPJPE-path~(mm) &165.4 &160.2 &157.7 &{156.6} &155.6 &155.8 &155.4 \\
        VRAM~(MB) &3560 &3700 &4358 &{5246} &7286 &11916 &20876 \\
        Speed~(sample/s) &99.8 &96.1 &88.5 &{82.4} &64.2 &42.2 &24.1 \\
        \hline
    \end{tabular}
    }
    \vspace{-0.3em}
    \caption{
        SIF3D performance with various scene point cloud sizes.
    }
    \label{t4}
  \vspace{-0.7em}
\end{table}

\tb{(6) Scene point cloud size.}
The original size of scene point clouds from LiDAR sensors typically surpasses 500K \cite{zheng2022gimo}.
To optimize the use of scene point clouds, we initially down-sample them.
However, this approach risks omitting critical scene details.
To strike a balance between computational overhead and performance, 
we conduct experiments by varying the scene point cloud sizes to assess SIF3D's performance.
Table \ref{t4} reveals that the performance improves with the increase in scene point cloud size until the size exceeds 4096.
While the performance continues to advance with the larger point cloud size, 
a critical issue emerges in terms of computational overhead and memory consumption
(The VRAM and speed are measured on a NVIDIA RTX3090 GPU during training).
Therefore, in SIF3D, we opt for a trade-off point cloud size of 4096.

\begin{table}[t!]
    \vspace{-0.0em}
    \renewcommand\arraystretch{1}
    \centering
    \footnotesize
    \setlength{\tabcolsep}{1.4mm}{
    \begin{tabular}{c|ccccc}
        Aggregate Method &\tb{Last} &Mean &Max &Conv &Transformer \\
        \hline
        Traj-path~(mm) &\tb{594} &595 &597 &602 &605 \\
        MPJPE-path~(mm) &\tb{157.1} &157.7 &159.2 &161.3 &160.8 \\
        \hline
    \end{tabular}
    }
    \vspace{-0.3em}
    \caption{
    Performance of SIF3D with various temporal aggregators.
    }
    \label{t5}
  \vspace{-1.2em}
\end{table}

\tb{(7) Motion aggregator in TIA.}
In TIA, we consolidate the motion sequence into a unified motion embedding using a temporal motion aggregator. 
To investigate the influence of various action sequence aggregators on performance, 
we conduct experiments involving 5 aggregation strategies:
(1)~Last, capturing the last motion embedding in the sequence.
(2)~Mean, averaging across the entire motion sequence.
(3)~Max, employing max-pooling throughout the entire motion sequence.
(4)~Conv, applying a three-layer convolutional network to downsample the sequence.
(5)~Transformer, introducing a single-layer transformer decoder into the aggregation process.

The results in Table \ref{t5} reveal that the performance of SIF3D is not significantly affected by the choice of motion aggregator, 
with the first aggregator (Last) demonstrating the best performance. 
However, the two weighted methods (Conv and Transformer) surprisingly yield inferior results. 
We attribute this counterintuitive finding to the limited size of the dataset and the 
insufficient supervision for the aggregated global motion embedding, 
making it challenging for the weighted methods to learn effectively.

\vspace{-0.0em}
\section{Conclusion}
\vspace{-0.3em}
\label{sec:5}
This work presents a pioneering multi-modal sense-informed framework, SIF3D,
designed for human motion prediction within real-world 3D scenes.
By incorporating external 3D scene points and internal human gaze, 
SIF3D acquires the ability to perceive the scene and comprehend human intention using SCA and TIA. 
In evaluations conducted on two datasets of GIMO and GTA-1M, SIF3D achieves an
average destination trajectory deviation of 666mm and 836mm on each dataset,
and a 195.7mm and 227.7mm average destination MPJPE,
establishing a new state-of-the-art performance both in global trajectory and local pose prediction.
Consequently, our findings underscore the significance of scene and gaze information in scene-based motion prediction. 
Furthermore, we posit that delving into high-fidelity diverse human motion generation in real-world 3D scenarios holds promise as a compelling avenue for future exploration.

\section*{Acknowledgements}
\vspace{-0.3em}
This work was supported in part by the National Key Research and Development Program of China (2022YFC3602601),
in part by the Natural Science Foundation of Jiangsu Province (BK20220939),
and in part by the National Natural Science Foundation of China (62306141).

{
    \small
    \bibliographystyle{ieee_fullname}
    \bibliography{egbib}

\begin{thebibliography}{10}\itemsep=-1pt

\bibitem{aksan2020attention}
Emre Aksan, Manuel Kaufmann, Peng Cao, and Otmar Hilliges.
\newblock {A Spatio-temporal Transformer for 3D Human Motion Prediction}.
\newblock In {\em 3DV}, pages 565--574. IEEE, 2021.

\bibitem{aksan2021spatio}
Emre Aksan, Manuel Kaufmann, Peng Cao, and Otmar Hilliges.
\newblock A spatio-temporal transformer for 3d human motion prediction.
\newblock In {\em 3DV}, pages 565--574. IEEE, 2021.

\bibitem{altche2017lstm}
Florent Altch{\'e} and Arnaud de La~Fortelle.
\newblock An lstm network for highway trajectory prediction.
\newblock In {\em 2017 IEEE 20th international conference on intelligent transportation systems (ITSC)}, pages 353--359. IEEE, 2017.

\bibitem{barsoum2018hp}
Emad Barsoum, John Kender, and Zicheng Liu.
\newblock {HP-GAN: Probabilistic 3D Human Motion Prediction via GAN}.
\newblock In {\em CVPR}, pages 1418--1427, 2018.

\bibitem{cao2020long}
Zhe Cao, Hang Gao, Karttikeya Mangalam, Qi-Zhi Cai, Minh Vo, and Jitendra Malik.
\newblock Long-term human motion prediction with scene context.
\newblock In {\em Computer Vision--ECCV 2020: 16th European Conference, Glasgow, UK, August 23--28, 2020, Proceedings, Part I 16}, pages 387--404. Springer, 2020.

\bibitem{chen2019deep}
Jingdao Chen, Zsolt Kira, and Yong~K Cho.
\newblock Deep learning approach to point cloud scene understanding for automated scan to 3d reconstruction.
\newblock {\em Journal of Computing in Civil Engineering}, 33(4):04019027, 2019.

\bibitem{choy20194d}
Christopher Choy, JunYoung Gwak, and Silvio Savarese.
\newblock 4d spatio-temporal convnets: Minkowski convolutional neural networks.
\newblock In {\em Proceedings of the IEEE/CVF conference on computer vision and pattern recognition}, pages 3075--3084, 2019.

\bibitem{corona2020context}
Enric Corona, Albert Pumarola, Guillem Alenya, and Francesc Moreno-Noguer.
\newblock {Context-aware Human Motion Prediction}.
\newblock In {\em CVPR}, pages 6992--7001, 2020.

\bibitem{Cui_2021_CVPR}
Qiongjie Cui and Huaijiang Sun.
\newblock {Towards Accurate 3D Human Motion Prediction From Incomplete Observations}.
\newblock In {\em CVPR}, pages 4801--4810, June 2021.

\bibitem{cui2020learning}
Qiongjie Cui, Huaijiang Sun, and Fei Yang.
\newblock {Learning Dynamic Relationships for 3D Human Motion Prediction}.
\newblock In {\em CVPR}, pages 6519--6527, 2020.

\bibitem{dang2021msr}
Lingwei Dang, Yongwei Nie, Chengjiang Long, Qing Zhang, and Guiqing Li.
\newblock {MSR-GCN: Multi-Scale Residual Graph Convolution Networks for Human Motion Prediction}.
\newblock In {\em ICCV}, pages 11467--11476, 2021.

\bibitem{diller2022forecasting}
Christian Diller, Thomas Funkhouser, and Angela Dai.
\newblock Forecasting characteristic 3d poses of human actions.
\newblock In {\em Proceedings of the IEEE/CVF Conference on Computer Vision and Pattern Recognition}, pages 15914--15923, 2022.

\bibitem{dong2022cswin}
Xiaoyi Dong, Jianmin Bao, Dongdong Chen, Weiming Zhang, Nenghai Yu, Lu Yuan, Dong Chen, and Baining Guo.
\newblock Cswin transformer: A general vision transformer backbone with cross-shaped windows.
\newblock In {\em Proceedings of the IEEE/CVF Conference on Computer Vision and Pattern Recognition}, pages 12124--12134, 2022.

\bibitem{engelmann20203d}
Francis Engelmann, Martin Bokeloh, Alireza Fathi, Bastian Leibe, and Matthias Nie{\ss}ner.
\newblock 3d-mpa: Multi-proposal aggregation for 3d semantic instance segmentation.
\newblock In {\em Proceedings of the IEEE/CVF conference on computer vision and pattern recognition}, pages 9031--9040, 2020.

\bibitem{fragkiadaki2015recurrent}
Katerina Fragkiadaki, Sergey Levine, Panna Felsen, and Jitendra Malik.
\newblock {Recurrent Network Models for Human Dynamics}.
\newblock In {\em ICCV}, pages 4346--4354, 2015.

\bibitem{ghosh2023imos}
Anindita Ghosh, Rishabh Dabral, Vladislav Golyanik, Christian Theobalt, and Philipp Slusallek.
\newblock Imos: Intent-driven full-body motion synthesis for human-object interactions.
\newblock In {\em Computer Graphics Forum}, volume~42, pages 1--12. Wiley Online Library, 2023.

\bibitem{graham20183d}
Benjamin Graham, Martin Engelcke, and Laurens Van Der~Maaten.
\newblock 3d semantic segmentation with submanifold sparse convolutional networks.
\newblock In {\em Proceedings of the IEEE conference on computer vision and pattern recognition}, pages 9224--9232, 2018.

\bibitem{Gui2018AdversarialGH}
Liang-Yan Gui, Yu-Xiong Wang, Xiaodan Liang, and Jos{\'e} M.~F. Moura.
\newblock {Adversarial Geometry-Aware Human Motion Prediction}.
\newblock In {\em ECCV}, pages 786--803, 2018.

\bibitem{Gui2018TeachingRT}
Liang-Yan Gui, Kevin Zhang, Yu-Xiong Wang, Xiaodan Liang, Jos{\'e} M.~F. Moura, and Manuela Veloso.
\newblock {Teaching Robots to Predict Human Motion}.
\newblock {\em IROS}, pages 562--567, 2018.

\bibitem{Guo2019HumanMP}
Xiao Guo and Jongmoo Choi.
\newblock {Human Motion Prediction via Learning Local Structure Representations and Temporal Dependencies}.
\newblock In {\em AAAI}, pages 2580--2587, 2019.

\bibitem{gwak2020generative}
JunYoung Gwak, Christopher Choy, and Silvio Savarese.
\newblock Generative sparse detection networks for 3d single-shot object detection.
\newblock In {\em Computer Vision--ECCV 2020: 16th European Conference, Glasgow, UK, August 23--28, 2020, Proceedings, Part IV 16}, pages 297--313. Springer, 2020.

\bibitem{hassan2021stochastic}
Mohamed Hassan, Duygu Ceylan, Ruben Villegas, Jun Saito, Jimei Yang, Yi Zhou, and Michael~J Black.
\newblock Stochastic scene-aware motion prediction.
\newblock In {\em Proceedings of the IEEE/CVF International Conference on Computer Vision}, pages 11374--11384, 2021.

\bibitem{hou20193d}
Ji Hou, Angela Dai, and Matthias Nie{\ss}ner.
\newblock 3d-sis: 3d semantic instance segmentation of rgb-d scans.
\newblock In {\em Proceedings of the IEEE/CVF conference on computer vision and pattern recognition}, pages 4421--4430, 2019.

\bibitem{hou2020revealnet}
Ji Hou, Angela Dai, and Matthias Nie{\ss}ner.
\newblock Revealnet: Seeing behind objects in rgb-d scans.
\newblock In {\em Proceedings of the IEEE/CVF Conference on Computer Vision and Pattern Recognition}, pages 2098--2107, 2020.

\bibitem{huang2023diffusion}
Siyuan Huang, Zan Wang, Puhao Li, Baoxiong Jia, Tengyu Liu, Yixin Zhu, Wei Liang, and Song-Chun Zhu.
\newblock Diffusion-based generation, optimization, and planning in 3d scenes.
\newblock In {\em Proceedings of the IEEE/CVF Conference on Computer Vision and Pattern Recognition}, pages 16750--16761, 2023.

\bibitem{ionescu2013human3}
Catalin Ionescu, Dragos Papava, Vlad Olaru, and Cristian Sminchisescu.
\newblock Human3. 6m: Large scale datasets and predictive methods for 3d human sensing in natural environments.
\newblock {\em IEEE transactions on pattern analysis and machine intelligence}, 36(7):1325--1339, 2013.

\bibitem{Jain2016StructuralRNNDL}
Ashesh Jain, Amir~Roshan Zamir, Silvio Savarese, and Ashutosh Saxena.
\newblock {Structural-RNN: Deep Learning on Spatio-Temporal Graphs}.
\newblock In {\em CVPR}, pages 5308--5317, 2016.

\bibitem{jiang2020end}
Haiyong Jiang, Feilong Yan, Jianfei Cai, Jianmin Zheng, and Jun Xiao.
\newblock End-to-end 3d point cloud instance segmentation without detection.
\newblock In {\em Proceedings of the IEEE/CVF Conference on Computer Vision and Pattern Recognition}, pages 12796--12805, 2020.

\bibitem{kipf2016semi}
Thomas~N Kipf and Max Welling.
\newblock {Semi-supervised Classification with Graph Convolutional Networks}.
\newblock {\em arXiv preprint arXiv:1609.02907}, 2016.

\bibitem{kundu2018bihmp}
Jogendra~Nath Kundu, Maharshi Gor, and R.~Venkatesh Babu.
\newblock {BiHMP-GAN: Bidirectional 3D Human Motion Prediction GAN}.
\newblock In {\em AAAI}, volume~33, pages 8553--8560, 2019.

\bibitem{li2020multitask}
Bin Li, Jian Tian, Zhongfei Zhang, Hailin Feng, and Xi Li.
\newblock {Multitask Non-Autoregressive Model for Human Motion Prediction}.
\newblock {\em IEEE Transactions on Image Processing}, 2020.

\bibitem{li2018convolutional}
Chen Li, Zhen Zhang, Wee Sun~Lee, and Gim Hee~Lee.
\newblock {Convolutional Sequence to Sequence Model for Human Dynamics}.
\newblock In {\em CVPR}, pages 5226--5234, 2018.

\bibitem{li2022skeleton}
Maosen Li, Siheng Chen, Zijing Zhang, Lingxi Xie, Qi Tian, and Ya Zhang.
\newblock {Skeleton-Parted Graph Scattering Networks for 3D Human Motion Prediction}.
\newblock In {\em ECCV}, pages 18--36. Springer, 2022.

\bibitem{li2020dynamic}
Maosen Li, Siheng Chen, Yangheng Zhao, Ya Zhang, Yanfeng Wang, and Qi Tian.
\newblock {Dynamic Multiscale Graph Neural Networks for 3D Skeleton Based Human Motion Prediction}.
\newblock In {\em CVPR}, pages 214--223, 2020.

\bibitem{liu2019flownet3d}
Xingyu Liu, Charles~R Qi, and Leonidas~J Guibas.
\newblock Flownet3d: Learning scene flow in 3d point clouds.
\newblock In {\em Proceedings of the IEEE/CVF conference on computer vision and pattern recognition}, pages 529--537, 2019.

\bibitem{liu2021multimodal}
Yicheng Liu, Jinghuai Zhang, Liangji Fang, Qinhong Jiang, and Bolei Zhou.
\newblock Multimodal motion prediction with stacked transformers.
\newblock In {\em Proceedings of the IEEE/CVF Conference on Computer Vision and Pattern Recognition}, pages 7577--7586, 2021.

\bibitem{liu2022swin}
Ze Liu, Han Hu, Yutong Lin, Zhuliang Yao, Zhenda Xie, Yixuan Wei, Jia Ning, Yue Cao, Zheng Zhang, Li Dong, et~al.
\newblock Swin transformer v2: Scaling up capacity and resolution.
\newblock In {\em Proceedings of the IEEE/CVF conference on computer vision and pattern recognition}, pages 12009--12019, 2022.

\bibitem{liu2021swin}
Ze Liu, Yutong Lin, Yue Cao, Han Hu, Yixuan Wei, Zheng Zhang, Stephen Lin, and Baining Guo.
\newblock Swin transformer: Hierarchical vision transformer using shifted windows.
\newblock In {\em Proceedings of the IEEE/CVF international conference on computer vision}, pages 10012--10022, 2021.

\bibitem{liu2021aggregated}
Zhenguang Liu, Kedi Lyu, Shuang Wu, Haipeng Chen, Yanbin Hao, and Shouling Ji.
\newblock {Aggregated Multi-GANs for Controlled 3D Human Motion Prediction}.
\newblock In {\em AAAI}, volume~35, pages 2225--2232, 2021.

\bibitem{ma2022progressively}
Tiezheng Ma, Yongwei Nie, Chengjiang Long, Qing Zhang, and Guiqing Li.
\newblock Progressively generating better initial guesses towards next stages for high-quality human motion prediction.
\newblock In {\em Proceedings of the IEEE/CVF Conference on Computer Vision and Pattern Recognition}, pages 6437--6446, 2022.

\bibitem{ma2022progy}
Tiezheng Ma, Yongwei Nie, Chengjiang Long, Qing Zhang, and Guiqing Li.
\newblock {Progressively Generating Better Initial Guesses Towards Next Stages for High-Quality Human Motion Prediction}.
\newblock In {\em CVPR}, pages 6437--6446, 2022.

\bibitem{mao2019learning}
Wei Mao, Miaomiao Liu, Mathieu Salzmann, and Hongdong Li.
\newblock {Learning Trajectory Dependencies for Human Motion Prediction}.
\newblock In {\em ICCV}, pages 9489--9497, 2019.

\bibitem{martinez2017human}
Julieta Martinez, Michael~J Black, and Javier Romero.
\newblock {On Human Motion Prediction using Recurrent Neural Networks}.
\newblock In {\em CVPR}, pages 2891--2900, 2017.

\bibitem{martinez2021pose}
Angel Mart{\'\i}nez-Gonz{\'a}lez, Michael Villamizar, and Jean-Marc Odobez.
\newblock Pose transformers (potr): Human motion prediction with non-autoregressive transformers.
\newblock In {\em Proceedings of the IEEE/CVF International Conference on Computer Vision}, pages 2276--2284, 2021.

\bibitem{maturana2015voxnet}
Daniel Maturana and Sebastian Scherer.
\newblock Voxnet: A 3d convolutional neural network for real-time object recognition.
\newblock In {\em 2015 IEEE/RSJ international conference on intelligent robots and systems (IROS)}, pages 922--928. IEEE, 2015.

\bibitem{nikhil2018convolutional}
Nishant Nikhil and Brendan Tran~Morris.
\newblock Convolutional neural network for trajectory prediction.
\newblock In {\em Proceedings of the European Conference on Computer Vision (ECCV) Workshops}, pages 0--0, 2018.

\bibitem{pavlakos2019expressive}
Georgios Pavlakos, Vasileios Choutas, Nima Ghorbani, Timo Bolkart, Ahmed~AA Osman, Dimitrios Tzionas, and Michael~J Black.
\newblock Expressive body capture: 3d hands, face, and body from a single image.
\newblock In {\em Proceedings of the IEEE/CVF conference on computer vision and pattern recognition}, pages 10975--10985, 2019.

\bibitem{piergiovanni2020adversarial}
AJ Piergiovanni, Anelia Angelova, Alexander Toshev, and Michael~S Ryoo.
\newblock {Adversarial Generative Grammars for Human Activity Prediction}.
\newblock In {\em ECCV}, pages 507--523. Springer, 2020.

\bibitem{qi2020imvotenet}
Charles~R Qi, Xinlei Chen, Or Litany, and Leonidas~J Guibas.
\newblock Imvotenet: Boosting 3d object detection in point clouds with image votes.
\newblock In {\em Proceedings of the IEEE/CVF conference on computer vision and pattern recognition}, pages 4404--4413, 2020.

\bibitem{qi2019deep}
Charles~R Qi, Or Litany, Kaiming He, and Leonidas~J Guibas.
\newblock Deep hough voting for 3d object detection in point clouds.
\newblock In {\em proceedings of the IEEE/CVF International Conference on Computer Vision}, pages 9277--9286, 2019.

\bibitem{qi2017pointnet}
Charles~R Qi, Hao Su, Kaichun Mo, and Leonidas~J Guibas.
\newblock Pointnet: Deep learning on point sets for 3d classification and segmentation.
\newblock In {\em Proceedings of the IEEE conference on computer vision and pattern recognition}, pages 652--660, 2017.

\bibitem{qi2016volumetric}
Charles~R Qi, Hao Su, Matthias Nie{\ss}ner, Angela Dai, Mengyuan Yan, and Leonidas~J Guibas.
\newblock Volumetric and multi-view cnns for object classification on 3d data.
\newblock In {\em Proceedings of the IEEE conference on computer vision and pattern recognition}, pages 5648--5656, 2016.

\bibitem{qi2017pointnet++}
Charles~Ruizhongtai Qi, Li Yi, Hao Su, and Leonidas~J Guibas.
\newblock Pointnet++: Deep hierarchical feature learning on point sets in a metric space.
\newblock {\em Advances in neural information processing systems}, 30, 2017.

\bibitem{radford2021learning}
Alec Radford, Jong~Wook Kim, Chris Hallacy, Aditya Ramesh, Gabriel Goh, Sandhini Agarwal, Girish Sastry, Amanda Askell, Pamela Mishkin, Jack Clark, et~al.
\newblock Learning transferable visual models from natural language supervision.
\newblock In {\em International conference on machine learning}, pages 8748--8763. PMLR, 2021.

\bibitem{radwan2020multimodal}
Noha Radwan, Wolfram Burgard, and Abhinav Valada.
\newblock Multimodal interaction-aware motion prediction for autonomous street crossing.
\newblock {\em The International Journal of Robotics Research}, 39(13):1567--1598, 2020.

\bibitem{Ruiz2018HumanMP}
Alejandro~Hernandez Ruiz, Juergen Gall, and Francesc Moreno-Noguer.
\newblock {Human Motion Prediction via Spatio-Temporal Inpainting}.
\newblock In {\em CVPR}, pages 7134--7143, 2018.

\bibitem{rusu2009close}
Radu~Bogdan Rusu, Nico Blodow, Zoltan~Csaba Marton, and Michael Beetz.
\newblock Close-range scene segmentation and reconstruction of 3d point cloud maps for mobile manipulation in domestic environments.
\newblock In {\em 2009 IEEE/RSJ International Conference on Intelligent Robots and Systems}, pages 1--6. IEEE, 2009.

\bibitem{su2015multi}
Hang Su, Subhransu Maji, Evangelos Kalogerakis, and Erik Learned-Miller.
\newblock Multi-view convolutional neural networks for 3d shape recognition.
\newblock In {\em Proceedings of the IEEE international conference on computer vision}, pages 945--953, 2015.

\bibitem{su2022crossmodal}
Zhaoxin Su, Gang Huang, Sanyuan Zhang, and Wei Hua.
\newblock Crossmodal transformer based generative framework for pedestrian trajectory prediction.
\newblock In {\em 2022 International Conference on Robotics and Automation (ICRA)}, pages 2337--2343. IEEE, 2022.

\bibitem{taheri2020grab}
Omid Taheri, Nima Ghorbani, Michael~J Black, and Dimitrios Tzionas.
\newblock Grab: A dataset of whole-body human grasping of objects.
\newblock In {\em Computer Vision--ECCV 2020: 16th European Conference, Glasgow, UK, August 23--28, 2020, Proceedings, Part IV 16}, pages 581--600. Springer, 2020.

\bibitem{Tang2018LongTermHM}
Yongyi Tang, Lin Ma, Wei Liu, and Wei-Shi Zheng.
\newblock {Long-Term Human Motion Prediction by Modeling Motion Context and Enhancing Motion Dynamics}.
\newblock In {\em IJCAI}, 2018.

\bibitem{thomas2019kpconv}
Hugues Thomas, Charles~R Qi, Jean-Emmanuel Deschaud, Beatriz Marcotegui, Fran{\c{c}}ois Goulette, and Leonidas~J Guibas.
\newblock Kpconv: Flexible and deformable convolution for point clouds.
\newblock In {\em Proceedings of the IEEE/CVF international conference on computer vision}, pages 6411--6420, 2019.

\bibitem{trick2019multimodal}
Susanne Trick, Dorothea Koert, Jan Peters, and Constantin~A Rothkopf.
\newblock Multimodal uncertainty reduction for intention recognition in human-robot interaction.
\newblock In {\em 2019 IEEE/RSJ International Conference on Intelligent Robots and Systems (IROS)}, pages 7009--7016. IEEE, 2019.

\bibitem{vaswani2017attention}
Ashish Vaswani, Noam Shazeer, Niki Parmar, Jakob Uszkoreit, Llion Jones, Aidan~N Gomez, {\L}ukasz Kaiser, and Illia Polosukhin.
\newblock {Attention is All You Need}.
\newblock In {\em NeurIPS}, pages 5998--6008, 2017.

\bibitem{wang2021pvred}
Hongsong Wang, Jian Dong, Bin Cheng, and Jiashi Feng.
\newblock {PVRED: A Position-Velocity Recurrent Encoder-Decoder for Human Motion Prediction}.
\newblock {\em IEEE Transactions on Image Processing}, 30:6096--6106, 2021.

\bibitem{wang2022towards}
Jingbo Wang, Yu Rong, Jingyuan Liu, Sijie Yan, Dahua Lin, and Bo Dai.
\newblock Towards diverse and natural scene-aware 3d human motion synthesis.
\newblock In {\em Proceedings of the IEEE/CVF Conference on Computer Vision and Pattern Recognition}, pages 20460--20469, 2022.

\bibitem{wang2021scene}
Jingbo Wang, Sijie Yan, Bo Dai, and Dahua Lin.
\newblock Scene-aware generative network for human motion synthesis.
\newblock In {\em Proceedings of the IEEE/CVF Conference on Computer Vision and Pattern Recognition}, pages 12206--12215, 2021.

\bibitem{wang2022humanise}
Zan Wang, Yixin Chen, Tengyu Liu, Yixin Zhu, Wei Liang, and Siyuan Huang.
\newblock Humanise: Language-conditioned human motion generation in 3d scenes.
\newblock {\em Advances in Neural Information Processing Systems}, 35:14959--14971, 2022.

\bibitem{wu20153d}
Zhirong Wu, Shuran Song, Aditya Khosla, Fisher Yu, Linguang Zhang, Xiaoou Tang, and Jianxiong Xiao.
\newblock 3d shapenets: A deep representation for volumetric shapes.
\newblock In {\em Proceedings of the IEEE conference on computer vision and pattern recognition}, pages 1912--1920, 2015.

\bibitem{xu2023auxiliary}
Chenxin Xu, Robby~T Tan, Yuhong Tan, Siheng Chen, Xinchao Wang, and Yanfeng Wang.
\newblock Auxiliary tasks benefit 3d skeleton-based human motion prediction.
\newblock In {\em Proceedings of the IEEE/CVF International Conference on Computer Vision}, pages 9509--9520, 2023.

\bibitem{yang2016automatic}
Bisheng Yang, Zhen Dong, Fuxun Liang, and Yuan Liu.
\newblock Automatic registration of large-scale urban scene point clouds based on semantic feature points.
\newblock {\em ISPRS Journal of Photogrammetry and Remote Sensing}, 113:43--58, 2016.

\bibitem{yu2015human}
Zhibin Yu, Sangwook Kim, Rammohan Mallipeddi, and Minho Lee.
\newblock Human intention understanding based on object affordance and action classification.
\newblock In {\em 2015 International Joint Conference on Neural Networks (IJCNN)}, pages 1--6. IEEE, 2015.

\bibitem{zheng2022gimo}
Yang Zheng, Yanchao Yang, Kaichun Mo, Jiaman Li, Tao Yu, Yebin Liu, C~Karen Liu, and Leonidas~J Guibas.
\newblock Gimo: Gaze-informed human motion prediction in context.
\newblock In {\em European Conference on Computer Vision}, pages 676--694. Springer, 2022.

\bibitem{zhong2023rspt}
Fangwei Zhong, Xiao Bi, Yudi Zhang, Wei Zhang, and Yizhou Wang.
\newblock Rspt: reconstruct surroundings and predict trajectory for generalizable active object tracking.
\newblock In {\em Proceedings of the AAAI Conference on Artificial Intelligence}, volume~37, pages 3705--3714, 2023.

\end{thebibliography}
}

\end{document}